\documentclass[11pt,a4paper]{article}

\usepackage[margin=1in]{geometry}
\usepackage[utf8]{inputenc}
\usepackage[T1]{fontenc}
\usepackage{mathpazo}
\usepackage{amsmath,amssymb}
\usepackage{graphicx}
\usepackage{booktabs}
\usepackage{tabularx}
\usepackage{array}
\usepackage{multirow}
\usepackage{xcolor}
\usepackage{float}
\usepackage{enumitem}
\usepackage{url}
\usepackage[colorlinks=true,linkcolor=blue!60!black,citecolor=blue!60!black,urlcolor=blue!70!black]{hyperref}
\usepackage{setspace}
\usepackage{titlesec}
\usepackage{caption}

\titleformat{\section}{\large\bfseries}{\thesection.}{0.5em}{}
\titleformat{\subsection}{\normalsize\bfseries}{\thesubsection.}{0.5em}{}
\begin{document}

% ---- TITLE ----
\begin{center}
{\LARGE\bfseries Governing the Agentic Enterprise: A Governance Maturity Model for Managing AI Agent Sprawl in Business Operations}\\[14pt]
{\large Vivek Acharya}\\[4pt]
{\small Independent Researcher, Dallas, TX, USA}\\
{\small Email: vacharya@bu.edu ~$\cdot$~ ORCID: \href{https://orcid.org/0009-0002-0860-9462}{0009-0002-0860-9462}}\\[12pt]
\rule{0.6\textwidth}{0.5pt}
\end{center}
\vspace{8pt}

% ---- ABSTRACT ----
\noindent\textbf{Abstract:}
The rapid adoption of agentic AI in enterprise business operations---autonomous systems capable of planning, reasoning, and executing multi-step workflows---has created an urgent governance crisis. Organizations face uncontrolled agent sprawl: the proliferation of redundant, ungoverned, and conflicting AI agents across business functions. Industry surveys report that only 21\% of enterprises have mature governance models for autonomous agents, while 40\% of agentic AI projects are projected to fail by 2027 due to inadequate governance and risk controls. Despite growing acknowledgment of this challenge, academic literature lacks a formal, empirically validated governance maturity model connecting governance capability to measurable business outcomes. This paper introduces the Agentic AI Governance Maturity Model (AAGMM), a five-level framework spanning 12 governance domains, grounded in NIST AI RMF and ISO/IEC 42001 standards. We additionally propose a novel taxonomy of agent sprawl patterns---functional duplication, shadow agents, orphaned agents, permission creep, and unmonitored delegation chains---each linked to quantifiable business cost models. The framework is validated through 750 simulation runs across five enterprise scenarios and five governance maturity levels, measuring business outcomes including cost containment, risk incident rates, operational efficiency, and decision quality. Results demonstrate statistically significant differences ($p < 0.001$, large effect sizes $d > 2.0$) between all governance maturity levels, with Level~4--5 organizations achieving 94.3\% lower sprawl indices, 96.4\% fewer risk incidents, and 32.6\% higher effective task completion rates compared to Level~1. The AAGMM provides practitioners with an actionable roadmap for governing autonomous AI agents while maximizing business returns.

\vspace{6pt}
\vspace{12pt}

% ============================================================
\section{Introduction}\label{sec:introduction}

The emergence of agentic artificial intelligence (AI) represents a paradigm shift in enterprise technology. Unlike traditional AI systems that respond to queries or classify inputs, agentic AI systems autonomously plan multi-step strategies, select and invoke tools, and execute complex business workflows with minimal human intervention~\cite{wang2024survey,acharya2025agentic}. These systems have rapidly moved beyond experimentation: by 2025, 79\% of organizations reported some level of agentic AI adoption, with the market projected to grow from \$8.5~billion in 2026 to \$45~billion by 2030~\cite{deloitte2026state,wef2026agentic}.

This adoption trajectory has generated immense business value. Google Cloud's 2025 ROI Report documented that 74\% of executives deploying AI agents achieved return on investment within the first year, with 39\% reporting doubled productivity in specific workflows~\cite{googlecloud2025roi}. Organizations across financial services, healthcare, retail, and manufacturing are deploying autonomous agents for customer service, fraud detection, supply chain optimization, and operational decision-making~\cite{deloitte2025emerging,kpmg2026q4pulse}.

However, this rapid adoption has simultaneously created a governance crisis that threatens to undermine the very value it generates. Three converging problems define this crisis:

\textbf{First, agent sprawl.} As low-code and no-code platforms make agent creation accessible to anyone in the organization, enterprises face the uncontrolled proliferation of redundant, fragmented, and ungoverned agents across teams and functions~\cite{googlecloud2026blueprint}. McKinsey identifies this as a risk analogous to the early days of robotic process automation: agents multiply across teams, duplicate efforts, and operate without oversight~\cite{mckinsey2025seizing}.

\textbf{Second, the governance gap.} Despite 74\% of companies planning agentic AI deployments, only 21\% of leaders surveyed have a mature governance model for autonomous agents~\cite{deloitte2026state}. Gartner projects that 40\% of agentic AI projects will fail by 2027 due to escalating costs, unclear business value, and inadequate risk controls~\cite{arcade2025trends}. The disconnect between adoption velocity and governance readiness is striking: 42\% of organizations report still developing their agentic strategy roadmap, with 35\% having no formal strategy at all~\cite{deloitte2025emerging}.

\textbf{Third, the ROI measurement vacuum.} While industry reports document impressive returns, they notably lack frameworks for how governance contributes to or detracts from these returns. Google Cloud's comprehensive ROI report, based on 3,466 executives, contains no discussion of agent governance models---how agents should be monitored, overridden, or audited~\cite{aigl2025roi}. Traditional ROI models are too narrow for agentic AI, focusing solely on direct cost savings while overlooking governance-dependent outcomes~\cite{moveworks2025roi}.

Academic research has begun addressing individual facets of this problem. Recent work includes the Enterprise Agentic Mesh (EAM) architecture for network-layer governance~\cite{gupta2025eam}, the Governance-as-a-Service (GaaS) framework for runtime compliance filtering~\cite{fernandez2025gaas}, comprehensive TRiSM reviews~\cite{ray2025trism}, and multi-expert analyses~\cite{crick2025agents}. However, none provide a holistic governance maturity model that (a)~defines progressive capability levels, (b)~maps controls to business outcomes, and (c)~empirically validates the governance--value relationship.

This paper addresses this gap with the Agentic AI Governance Maturity Model (AAGMM). Our contributions are fourfold:

\begin{itemize}[leftmargin=*, itemsep=2pt]
\item[\textbf{C1.}] The AAGMM framework: a five-level model spanning 12 governance domains, grounded in NIST AI RMF 1.0~\cite{nist2023rmf} and ISO/IEC 42001~\cite{iso42001}.
\item[\textbf{C2.}] A governance-to-ROI mapping connecting maturity levels to four business outcome dimensions.
\item[\textbf{C3.}] A novel taxonomy of five agent sprawl patterns with quantifiable cost models.
\item[\textbf{C4.}] Experimental validation through 750 simulation runs ($p < 0.001$ for all pairwise comparisons).
\end{itemize}

The paper is organized as follows: Section~\ref{sec:related} reviews related work; Section~\ref{sec:taxonomy} presents the sprawl taxonomy; Section~\ref{sec:framework} details the AAGMM; Section~\ref{sec:experimental} describes experimental design; Section~\ref{sec:results} presents results; Section~\ref{sec:discussion} discusses implications; Section~\ref{sec:conclusion} concludes.

% ============================================================
\section{Related Work}\label{sec:related}

\subsection{Agentic AI in Enterprise Business Operations}

Agentic AI represents a fundamental evolution from generative AI. While generative AI excels at content production, agentic AI systems combine LLMs with tool access, memory, and planning to autonomously pursue complex goals~\cite{wang2024survey,acharya2025agentic}. The 2025 MIT Sloan Management Review study found that agentic AI complicates traditional management boundaries: these systems plan, learn, and adapt as organizational actors~\cite{mitsloan2025emerging}.

Deloitte's 2026 report documents expanding use cases across customer support, supply chain, R\&D, and cybersecurity~\cite{deloitte2026state}. KPMG reported 67\% of leaders will maintain AI spending even during recession, with \$124M projected per organization~\cite{kpmg2026q4pulse}. Agentic AI is projected to add \$2.6--4.4T to global GDP by 2030~\cite{arcade2025trends}.

\subsection{AI Governance Frameworks}

NIST AI RMF 1.0 provides the most widely referenced structure (Govern, Map, Measure, Manage)~\cite{nist2023rmf}. NIST AI 600-1 extends this to generative AI~\cite{nist2024genai}. ISO/IEC 42001 establishes certifiable AI management requirements~\cite{iso42001}. The EU AI Act introduces risk-tiered obligations~\cite{euaiact2024}. However, these were designed for traditional deployments---not for environments with dozens of autonomous agents delegating tasks and making decisions without oversight.

\subsection{Maturity Models in Technology Governance}

CMMI defines five process maturity levels widely adopted in software engineering~\cite{cmmi2018}. COBIT provides IT governance maturity~\cite{cobit2019}. Recent AI maturity models~\cite{gartner2024aimaturity,microsoft2024aimaturity} address adoption readiness but not autonomous multi-agent governance specifically.

\subsection{Agent Governance and Multi-Agent Systems}

The EAM proposed a Semantic Control Plane and Governance Sidecar~\cite{gupta2025eam}. GaaS introduced runtime compliance filtering with trust scores~\cite{fernandez2025gaas}. TRiSM reviews cataloged trust/risk/security challenges~\cite{ray2025trism}. Manager Agent research formalized orchestration as a POSG~\cite{chen2025manageragent}. The LOKA Protocol proposed decentralized governance~\cite{ranjan2025loka}. OpenAI published agentic governance practices~\cite{openai2025practices}.

Despite this work, no framework provides a prescriptive maturity model mapping governance capabilities to business outcomes with empirical validation. The AAGMM fills this gap.

% ============================================================
\section{Agent Sprawl Taxonomy}\label{sec:taxonomy}

\subsection{Taxonomy of Sprawl Patterns}

Table~\ref{tab:sprawl-taxonomy} presents five sprawl patterns with business impacts.

\begin{table}[H]
\caption{Agent Sprawl Taxonomy.}\label{tab:sprawl-taxonomy}
\centering\small
\begin{tabularx}{\textwidth}{c l X X}
\toprule
\textbf{ID} & \textbf{Pattern} & \textbf{Definition} & \textbf{Business Impact} \\
\midrule
SP-01 & Functional Duplication & Multiple agents performing identical tasks across teams & Wasted compute (2--5$\times$), inconsistent outputs \\[4pt]
SP-02 & Shadow Agents & Agents deployed outside IT governance without registration & Security blind spots, compliance violations \\[4pt]
SP-03 & Orphaned Agents & Agents whose purpose expired but continue operating & Ongoing costs for zero value, security surface expansion \\[4pt]
SP-04 & Permission Creep & Agents accumulating permissions beyond original scope & Privilege escalation, regulatory non-compliance \\[4pt]
SP-05 & Unmonitored Delegation & Multi-agent chains without visibility or authorization tracking & Loss of accountability, cascading failures \\
\bottomrule
\end{tabularx}
\end{table}

\subsection{Sprawl Cost Model}

We define total sprawl cost as:
\begin{equation}\label{eq:sprawl-cost}
C_{\text{sprawl}} = C_{\text{redundancy}} + C_{\text{security}} + C_{\text{compliance}} + C_{\text{operational}} + C_{\text{opportunity}}
\end{equation}

\noindent This model demonstrates that governance is investment, not overhead: there exists a maturity level at which net business value is maximized.

% ============================================================
\section{The AAGMM Framework}\label{sec:framework}

The AAGMM defines five progressive capability levels across 12 governance domains, grounded in NIST AI RMF 1.0 and ISO/IEC 42001.

\subsection{Design Principles}

Four principles guide the AAGMM: \textbf{Progressive capability} (each level builds on the previous); \textbf{Business-outcome orientation} (levels defined by outcomes, not just controls); \textbf{Domain completeness} (12 domains cover the full agent lifecycle); \textbf{Standards alignment} (every domain maps to NIST/ISO clauses).

\subsection{Five Maturity Levels}

\begin{table}[H]
\caption{AAGMM Maturity Levels.}\label{tab:maturity-levels}
\centering\small
\begin{tabularx}{\textwidth}{c l X X X}
\toprule
\textbf{Lv.} & \textbf{Name} & \textbf{Key Characteristics} & \textbf{Active Controls} & \textbf{Expected Outcomes} \\
\midrule
L1 & Ad-hoc & No formal governance; agents deployed without coordination & None & High sprawl; uncontrolled costs \\[4pt]
L2 & Reactive & Basic registry; incident-triggered governance & Registry, incident response & Reduced blind spots \\[4pt]
L3 & Defined & Formal policies; centralized catalog; RBAC; HITL & Registry, IAM, observability, policy, HITL, audit, risk class. & Sprawl containment; measurable compliance \\[4pt]
L4 & Managed & Quantitative metrics; automated enforcement & All L3 + auto policy, sprawl detection, lifecycle & Proactive risk reduction \\[4pt]
L5 & Optimized & Self-improving; predictive sprawl; governance-as-code & All L4 + continuous improvement, predictive, GaC & Max value; min overhead \\
\bottomrule
\end{tabularx}
\end{table}

\subsection{Twelve Governance Domains}

\begin{table}[H]
\caption{Governance Domain Catalog with Standards Mapping.}\label{tab:domains}
\centering\small
\begin{tabularx}{\textwidth}{l l l l X}
\toprule
\textbf{ID} & \textbf{Domain} & \textbf{NIST RMF} & \textbf{ISO 42001} & \textbf{Sprawl Pattern} \\
\midrule
GD-01 & Agent Lifecycle Mgmt. & GOV 1.1, 1.3 & 6.1, 8.1 & SP-03 \\
GD-02 & Identity \& Access Ctrl. & GOV 1.5, MNG 2.3 & 9.2, 9.4 & SP-04 \\
GD-03 & Observability & MEA 2.1, 2.6 & 9.1 & SP-02, SP-05 \\
GD-04 & Policy Enforcement & GOV 1.2, MNG 4.1 & 5.2, 8.2 & All \\
GD-05 & Data Governance & MAP 3.1, GOV 1.7 & A.8 & SP-04 \\
GD-06 & Human Oversight & GOV 1.4, MNG 1.3 & 6.2 & SP-05 \\
GD-07 & Sprawl Containment & GOV 6.1, MAP 1.5 & 8.1, 10.1 & SP-01, SP-02 \\
GD-08 & Audit \& Compliance & MEA 2.8, MNG 4.2 & 9.2, 10.2 & All \\
GD-09 & Inter-Agent Coord. & MAP 3.4, MNG 2.1 & 8.2 & SP-05 \\
GD-10 & Risk Classification & MAP 1.1, 1.2 & 6.1.2 & SP-04 \\
GD-11 & Incident Response & MNG 3.1, 4.1 & A.6.2.5 & All \\
GD-12 & Continuous Improvement & MEA 3.1, GOV 6.2 & 10.1, 10.2 & All \\
\bottomrule
\end{tabularx}
\end{table}

These organize into four groups: \textbf{Lifecycle \& Identity} (GD-01--02) for registration, versioning, and least-privilege access; \textbf{Runtime Governance} (GD-03--06) for observability, policy-as-code, data boundaries, and HITL breakpoints; \textbf{Portfolio Governance} (GD-07--10) for sprawl scanning, audit, delegation protocols, and risk tiers; \textbf{Improvement \& Resilience} (GD-11--12) for quarantine, rollback, and governance KPIs.

\subsection{Governance-to-ROI Mapping}

We define four outcome dimensions: \textbf{Cost Containment} via Sprawl Index $\text{SI}$ and Governance Cost Ratio $\text{GCR}$; \textbf{Risk Reduction} via Risk Incident Rate $\text{RIR}$ (per 1K actions); \textbf{Operational Efficiency} via Effective Task Completion Rate $\text{ETCR}$; \textbf{Decision Quality} via Delegation Safety Rate $\text{DSR}$.

The composite Net Business Value:
\begin{equation}\label{eq:nbv}
\text{NBV} = 0.30 \!\cdot\! \text{ETCR} + 0.20 \!\cdot\! (1 \!-\! \text{SI}) + 0.20 \!\cdot\! (1 \!-\! \tfrac{\text{RIR}}{100}) + 0.15 \!\cdot\! \text{DSR} + 0.15 \!\cdot\! (1 \!-\! \text{GCR})
\end{equation}

Weights from Deloitte~\cite{deloitte2026state} and KPMG~\cite{kpmg2026q4pulse} priority surveys.

% ============================================================
\section{Experimental Design}\label{sec:experimental}

\subsection{Simulation Framework}

A Python-based multi-agent simulation with deterministic seeding ($\text{seed}=42$). Code at \url{https://github.com/curiosityexplorer/agentic-governance-maturity}. Five business functions, six capabilities each (30 types). Task difficulty: 40\% simple, 35\% moderate, 20\% complex, 5\% critical.

\subsection{Governance Configurations}

Shadow probability decreases L1 (0.35) to L5 (0.02). Orphan rates L1 (0.15) to L5 (0.01). Controls affect success probabilities, violation rates, and delegation safety.

\subsection{Experimental Scenarios}

\begin{table}[H]
\caption{Experimental Scenarios.}\label{tab:scenarios}
\centering\small
\begin{tabularx}{\textwidth}{l X c c X}
\toprule
\textbf{Scenario} & \textbf{Description} & \textbf{Agents} & \textbf{Tasks} & \textbf{Primary Stress} \\
\midrule
S1: Greenfield & New deployment across 5 units & 30 & 1{,}000 & Initial sprawl \\
S2: Scaling & Expansion to larger fleet & 50 & 2{,}000 & Scalability \\
S3: Cross-Func. & Multi-dept.\ collaboration & 35 & 1{,}500 & Delegation chains \\
S4: Adversarial & Unauthorized actions & 30 & 1{,}000 & Security \\
S5: Optimization & Active tuning/retirement & 40 & 1{,}500 & Portfolio mgmt. \\
\bottomrule
\end{tabularx}
\end{table}

\subsection{Statistical Methodology}

750 total runs ($5 \times 5 \times 30$). Welch's $t$-test + Cohen's $d$. Significance: $p<0.05$ (*), $p<0.01$ (**), $p<0.001$ (***).

% ============================================================
\section{Results and Analysis}\label{sec:results}

\subsection{Cross-Level Summary}

\begin{table}[H]
\caption{Business Outcomes by Governance Level ($n=150$ per level).}\label{tab:crosslevel}
\centering\small
\begin{tabular}{l c c c c c c}
\toprule
\textbf{Level} & \textbf{Sprawl} & \textbf{Risk/1K} & \textbf{Compl.} & \textbf{Gov.\ Cost} & \textbf{Del.\ Safe} & \textbf{NBV} \\
\midrule
L1 Ad-hoc    & $.520\!\pm\!.020$ & $59.08\!\pm\!1.11$ & $.699\!\pm\!.003$ & $.020$ & $.602\!\pm\!.007$ & $.625\!\pm\!.006$ \\
L2 Reactive  & $.378\!\pm\!.017$ & $40.26\!\pm\!0.89$ & $.729\!\pm\!.003$ & $.060$ & $.600\!\pm\!.007$ & $.694\!\pm\!.004$ \\
L3 Defined   & $.254\!\pm\!.017$ & $13.42\!\pm\!0.51$ & $.863\!\pm\!.002$ & $.120$ & $.902\!\pm\!.005$ & $.849\!\pm\!.004$ \\
L4 Managed   & $.065\!\pm\!.005$ & $3.45\!\pm\!0.33$  & $.890\!\pm\!.002$ & $.180$ & $.934\!\pm\!.004$ & $.910\!\pm\!.003$ \\
L5 Optimized & $.028\!\pm\!.004$ & $2.05\!\pm\!0.24$  & $.930\!\pm\!.002$ & $.160$ & $.992\!\pm\!.002$ & $.944\!\pm\!.002$ \\
\bottomrule
\end{tabular}
\end{table}

\subsection{Key Findings}

Figure~\ref{fig:metrics-panel} visualizes the four key business outcome metrics across all governance maturity levels.

\begin{figure}[H]
\centering
\includegraphics[width=\textwidth]{fig1_metrics_panel.png}
\caption{Business outcome metrics by governance maturity level ($n = 150$ per level, error bars = 95\% CI). (a)~Sprawl Index decreases 94.6\% from L1 to L5; (b)~Risk incidents decrease 96.5\%; (c)~Effective task completion improves 33.0\%; (d)~Composite Net Business Value improves 51.0\%.}\label{fig:metrics-panel}
\end{figure}

\textbf{Finding 1: Governance maturity dramatically reduces sprawl.} SI decreased from 0.520 (L1) to 0.028 (L5)---94.6\% reduction. Largest drop: L3$\to$L4 (74.4\%), coinciding with automated sprawl detection (GD-07).

\textbf{Finding 2: Risk follows a power-law decline.} RIR dropped from 59.08 to 2.05 (96.5\%). Sharpest: L2$\to$L3 (66.7\%), when policy enforcement and access controls activate.

\textbf{Finding 3: Task completion improves monotonically.} ETCR: 0.699$\to$0.930 (+33.0\%). Governance cost at L5 (0.160) is \textit{lower} than L4 (0.180)---automation reduces overhead.

\textbf{Finding 4: Delegation safety step-functions at L3.} DSR jumps from $\sim$0.60 (L1--2) to 0.902 (L3) with inter-agent coordination protocols.

\textbf{Finding 5: L5 delivers an automation dividend.} GCR \textit{decreases} L4$\to$L5 while all outcomes improve.

\subsection{Statistical Significance}

\begin{table}[H]
\caption{Pairwise Comparisons: Net Business Value.}\label{tab:pairwise}
\centering\small
\begin{tabular}{l c c c c l}
\toprule
\textbf{Comparison} & $\boldsymbol{\Delta}$ \textbf{NBV} & $\boldsymbol{t}$ & $\boldsymbol{p}$ & \textbf{Cohen's} $\boldsymbol{d}$ & \textbf{Effect} \\
\midrule
L1 vs.\ L2 & $+0.069$ & 19.87  & $<0.001$ & 2.30  & Large \\
L1 vs.\ L3 & $+0.224$ & 64.92  & $<0.001$ & 7.50  & Very Large \\
L1 vs.\ L4 & $+0.286$ & 100.13 & $<0.001$ & 11.56 & Very Large \\
L1 vs.\ L5 & $+0.319$ & 112.35 & $<0.001$ & 12.97 & Very Large \\
L2 vs.\ L3 & $+0.155$ & 51.10  & $<0.001$ & 5.90  & Very Large \\
L3 vs.\ L4 & $+0.062$ & 26.75  & $<0.001$ & 3.09  & Large \\
L3 vs.\ L5 & $+0.096$ & 41.59  & $<0.001$ & 4.80  & Very Large \\
L4 vs.\ L5 & $+0.034$ & 27.33  & $<0.001$ & 3.16  & Large \\
\bottomrule
\end{tabular}
\end{table}

All $d > 2.0$, far exceeding the large-effect threshold of 0.8.

\subsection{Scenario-Specific Analysis}

\begin{table}[H]
\caption{NBV by Scenario and Level (mean, $n=30$).}\label{tab:scenario-nbv}
\centering\small
\begin{tabular}{l c c c c c}
\toprule
\textbf{Scenario} & \textbf{L1} & \textbf{L2} & \textbf{L3} & \textbf{L4} & \textbf{L5} \\
\midrule
S1: Greenfield   & 0.616 & 0.717 & 0.863 & 0.924 & 0.953 \\
S2: Scaling      & 0.579 & 0.652 & 0.796 & 0.904 & 0.927 \\
S3: Cross-Func.  & 0.644 & 0.698 & 0.832 & 0.915 & 0.958 \\
S4: Adversarial  & 0.664 & 0.666 & 0.856 & 0.912 & 0.954 \\
S5: Optimization & 0.620 & 0.684 & 0.843 & 0.898 & 0.941 \\
\bottomrule
\end{tabular}
\end{table}

\textbf{S2 (Scaling)} showed worst L1 performance (NBV 0.579, SI 0.746), confirming sprawl compounds without governance. \textbf{S4 (Adversarial)} revealed L2 provides virtually no security benefit (NBV 0.666 vs.\ 0.664); L3 achieves 0.856 (+28.9\%). \textbf{S5 (Optimization)} showed widest L4--L5 gap, validating the automation dividend.

\begin{figure}[H]
\centering
\includegraphics[width=0.92\textwidth]{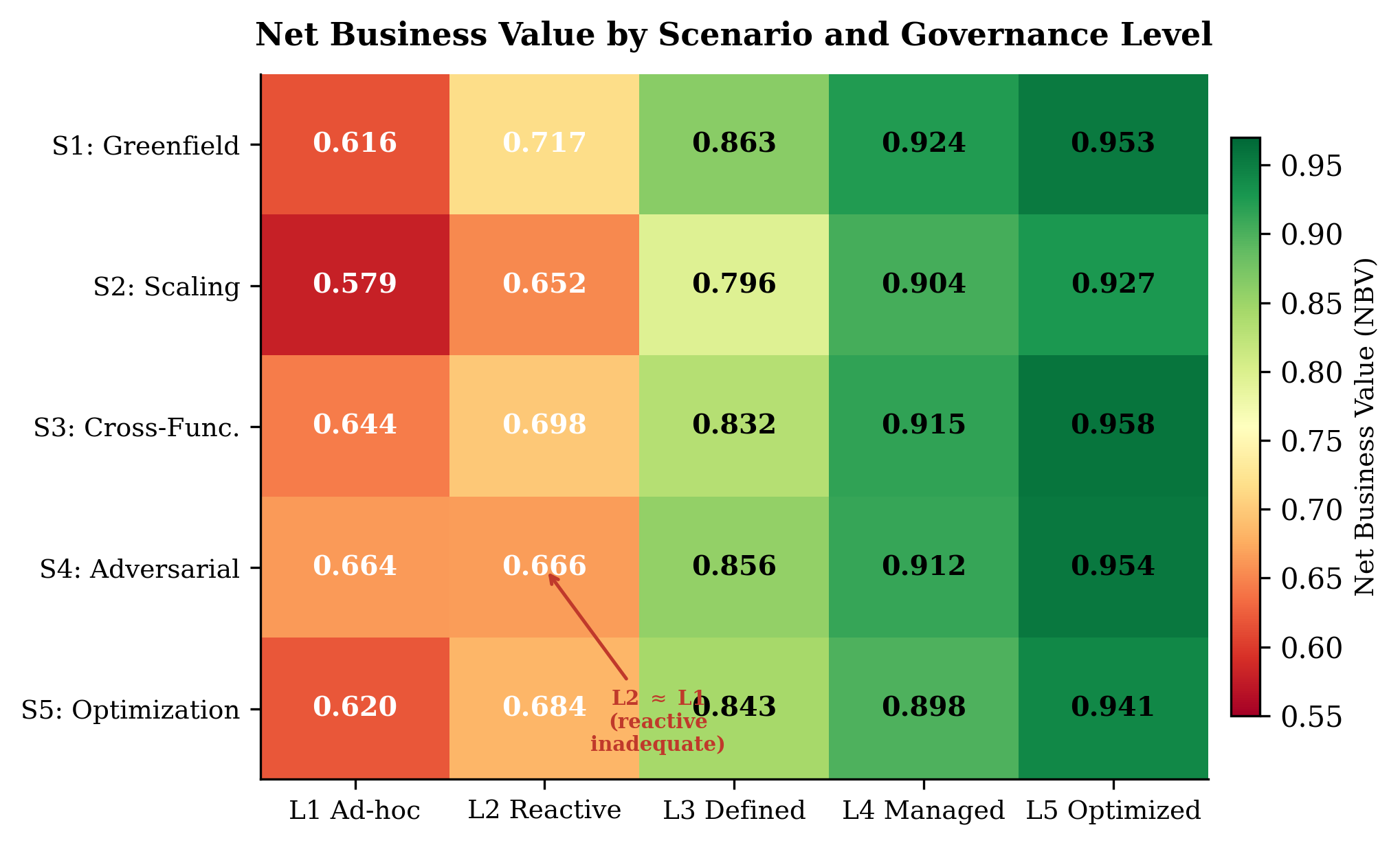}
\caption{Net Business Value heatmap across five experimental scenarios and five governance maturity levels ($n = 30$ per cell). The color gradient from red (low NBV) to green (high NBV) shows consistent governance benefits across all scenarios. Notable: in the Adversarial scenario (S4), L2 provides virtually no improvement over L1 (0.666 vs.\ 0.664), confirming that reactive governance is inadequate for security-sensitive deployments.}\label{fig:scenario-heatmap}
\end{figure}

Figure~\ref{fig:scenario-heatmap} visualizes these patterns as a heatmap, revealing the consistent improvement gradient from L1 (red) to L5 (green) across all scenarios, and highlighting the near-identical L1 and L2 values in the adversarial scenario.

% ============================================================
\section{Discussion}\label{sec:discussion}

\subsection{Practical Implications}

\textbf{Level~3 is the minimum viable governance standard} ($+0.155$ NBV, $d=5.90$ from L2). \textbf{Automated sprawl detection is the highest-leverage control} (74.4\% SI reduction at L4). \textbf{L5 delivers an automation dividend} (GCR decreases while outcomes improve). \textbf{Reactive governance is inadequate for security} (0.3\% NBV improvement L1$\to$L2 in adversarial scenario).

\subsection{Comparison with Existing Models}

The AAGMM differs from CMMI~\cite{cmmi2018} (no agent provisions), COBIT~\cite{cobit2019} (no real-time controls), and Gartner/Microsoft models~\cite{gartner2024aimaturity,microsoft2024aimaturity} (adoption-focused). Among academic work, EAM~\cite{gupta2025eam} addresses network governance and GaaS~\cite{fernandez2025gaas} focuses on output filtering---neither provides maturity levels linked to business outcomes. The AAGMM is the first to combine progressive levels, 12 standards-mapped domains, and empirical validation.

\subsection{Alignment with Industry Findings}

Deloitte's leadership-driven governance finding~\cite{deloitte2026state} aligns with monotonic NBV increase. KPMG's complexity-barrier finding~\cite{kpmg2026q4pulse} is addressed by our sprawl taxonomy. McKinsey's RPA analogy~\cite{mckinsey2025seizing} is validated: S2 L1 SI of 0.746 means three-quarters of agents are problematic.

\subsection{Limitations}

Validation uses simulation (not production deployments). Task/violation models are parameterized from reports. Governance costs are fixed per-level. NBV weights reflect one prioritization. Binary controls simplify continuous spectrums. Sensitivity analysis and case studies are future work.

\subsection{Ethical Considerations}

The AAGMM embeds ethics via Human Oversight (GD-06, HITL at L3+), Risk Classification (GD-10, EU AI Act-consistent tiers~\cite{euaiact2024}), and Audit \& Compliance (GD-08, decision traceability).

% ============================================================
\section{Conclusions and Future Work}\label{sec:conclusion}

This paper introduced the AAGMM, validated through 750 simulation runs. Key results: SI decreased 94.6\%; RIR decreased 96.5\%; ETCR improved 33.0\%; DSR improved 60.2\%$\to$99.2\%. All comparisons $p<0.001$, $d>2.0$.

Three practical findings: (1)~L3 is minimum viable governance; (2)~automated sprawl detection is highest-leverage; (3)~L5 delivers an automation dividend.

Future work: case study validation, self-assessment instrument, platform-specific guidance (LangChain, AutoGen, CrewAI), dynamic cost modeling, and MCP integration.

As agentic AI becomes the enterprise backbone, governance determines whether agent ecosystems generate compounding value or compounding risk. The AAGMM provides a structured, evidence-based path to the former.

% ---- STATEMENTS ----
\vspace{10pt}\noindent\rule{\textwidth}{0.4pt}\vspace{6pt}

\noindent\textbf{Author Contributions:} Conceptualization, methodology, software, validation, formal analysis, investigation, data curation, writing---original draft, writing---review \& editing, visualization: V.A.

\noindent\textbf{Funding:} This research received no external funding.

\noindent\textbf{Data Availability Statement:} Code, raw results, and aggregated data are openly available at \url{https://github.com/curiosityexplorer/agentic-governance-maturity} (seed = 42).

\noindent\textbf{Conflicts of Interest:} The author declares no conflicts of interest.

% ============================================================

\end{document}